\documentclass{article}

\usepackage{arxiv}

\usepackage[utf8]{inputenc} 
\usepackage[T1]{fontenc}    
\usepackage{hyperref}       
\usepackage{url}            
\usepackage{booktabs}       
\usepackage{amsfonts}       
\usepackage{nicefrac}       
\usepackage{microtype}      
\usepackage{lipsum}

\usepackage{color}
\usepackage{amsthm}
\usepackage{amsmath, amssymb}

\newtheorem{thm}{Theorem}
\usepackage{graphicx, hyperref}
\usepackage{tikz} 
\usetikzlibrary{matrix,chains,positioning,decorations.pathreplacing,arrows}
\usetikzlibrary{positioning,calc}
\usepackage[noend]{algpseudocode}
\usepackage{algorithm}
\usepackage{url}
\usepackage{cite}

\usepackage{authblk}

\title{A copula-based visualization technique for a neural network}
\author[ A, 1]{\textbf{Yusuke Kubo}}
\author[ B, 1, 2]{\textbf{ Yuto Komori}}
\author[ B]{\textbf{ Toyonobu Okuyama}}
\author[ B]{\textbf{ Hiroshi Tokieda}}
\affil[A]{Data Science Laboratories, NEC Corporation, Nakahara-ku, Kawasaki, Japan,}
\affil[B]{AI Research Center, National Institute of Advanced Industrial Science and Technology, Koto-ku, Tokyo, Japan}

\begin{document}
\maketitle
%
\footnotetext[1]{Both authors contributed equally to this study.}
\footnotetext[2]{Corresponding author for this study.}

\begin{abstract}
Interpretability of machine learning is defined as the extent to which
humans can comprehend the reason of a decision. However, a neural network
is not considered interpretable due to the ambiguity in its decision-making process.
Therefore, in this study, we propose a new algorithm that reveals which feature values 
the trained neural network considers important and which paths are mainly
traced in the process of decision-making.
In the proposed algorithm, the score estimated by the correlation
coefficients between the neural network layers
that can be calculated by applying the concept of a pair copula was defined.
We compared the estimated score with the feature importance values of
Random Forest,
which is sometimes regarded as a highly interpretable algorithm, in the experiment and
confirmed that the results were consistent with each other. 
This algorithm suggests an approach for compressing a neural network and its parameter tuning
because the algorithm identifies the paths that contribute
to the classification or prediction results.
\end{abstract}

\keywords{Neural Network \and Interpretability \and Copula}

\section{Introduction}
Interpretability of machine learning is being vigorously
discussed currently. Deep learning has an overwhelming
performance compared to other machine learning techniques
in the field of image recognition such as generic
object recognition. It is applied in a wide range
of fields, but is hesitated to be introduced under
the normal social conventions due to the difficulty of
interpretability in some cases. As for autonomous
driving technology, deep learning and
deep Q-network\cite{Mnih2015} 
bring the possibility of producing amazing
results\cite{Sallab2017}. 
However, the cause of accidents is unclear if traffic
accidents occur due to incorrect recognition. 
This problem also happens in the clinical practice.
When a doctor makes a diagnosis based on the machine
learning predictions, patients are not able to accept
the diagnosis if the predictions show the disease without 
appropriate reasons.
On top of that, security problems in machine
learning models against Adversarial
Attack\cite{Madry2018} 
break gradually to the surface.
It is recommended at the political level that the process
of machine learning decision-making is clear and that humans
should employ machine learning predictions
with a deep understanding of how they function\cite{Goodman2017}.
What is the interpretability or explainability of machine
learning after all?
A good reference\cite{Molnar2019} defined the interpretability of machine learning as 
``Interpretability is the degree to which
a human can understand the cause of a decision."
Based on this definition, the Linear Model and Decision
Tree can be considered as interpretable 
because in the Linear Model, it becomes clear which parameters are more important while in the Decision Tree, there is no ambiguity in the process of if-then-else statement.
A rule-based algorithm can also be termed interpretable.
However, a neural network is not interpretable based on the 
definition given above because the process of decision-making
is ambiguous although several novel approaches to clarify 
the process have been attempted.
Among them are the sensitivity analysis method that calculates the influence rate of the output by changing the input feature values
slightly and expresses it in a human-readable format\cite{Zeiler2014}; \cite{Smilkov2017},
the method that traces the network path from an output
to an input in reverse and identifies the effective input features\cite{Bach2015}, 
the method that replaces the trained model with the Linear
Model as an interpretable model and evaluates the
important input features\cite{Ribeiro2016},
the method that weighs the input feature values and predicts
the output using of them\cite{Vaswani2017}; \cite{Bahdanau2014} 
and the method that mimics the uninterpretable models by using
the interpretable models\cite{Bucila2006}; \cite{Ba2014}; \cite{Hinton2015}; \cite{Hendricks2016}.
In this study, we propose a new algorithm that enables us 
to clearly understand and visualize the decision-making 
process of a neural network.
The algorithm is realized by considering a neural
network as a graphical model and calculating the correlation
coefficients between its layers.
A copula is a general technique that is extremely useful 
for calculating the correlation coefficients, and therefore,
we first introduce its theoretical background in
the following section.
In Section \ref{sec:ALGORITHM}, the new algorithm is proposed 
using the concepts introduced in Section \ref{sec:PRELIMINARIES}.
In Section \ref{sec:EVALUATION}, an experiment using the
algorithm to determine the paths important in a neural network
for classification or prediction is evaluated using 
a well-known dataset.
Notably, the visualization result
is presented in Section \ref{sec:CCAL}, which presents
our main findings.
A comparison with Random Forest is described in
Section \ref{sec:COMPARISON}, which discusses the validity
of interpretability. Finally, the conclusion and future works
are summarized in Section \ref{sec:CONCLUSION}.

\section{PRELIMINARIES}
\label{sec:PRELIMINARIES}
In this section, the concepts of a copula, correlation, and
the relationship between them are introduced to define
the proposed algorithm.
\subsection{Copula}
\label{sec:COPULA}
Let $X_{1}, \ldots , X_{n}$ be random variables
and $x_{1}, \ldots , x_{n}$ be their values.
$F_{i}$ is the distribution function of the random
variable
$X_{i}$ for all $x_{i}$ in $\mathbb{R}$, therefore, 
$F_{i} = P(X_{i} \leq x_{i})$.
We focus on the multivariate cumulative joint distribution 
$F (x_{1}, \ldots ,x_{n})$ to consider the behavior
of multiple random variables.
There exists the following relationship between 
the joint distribution and the marginal distribution.
\begin{thm}\upshape\cite{Sklar1959} \itshape
\label{thm:DEFINITION_OF_COPULA}
Let $F_{1}(x_1),...,F_{n}(x_n)$ be the marginal distributions
of $F(x_1, \ldots ,x_n)$.
Then, there exists a functional C such that
\begin{equation*}
    F(x_1,...,x_n) = C(F_{1}(x_{1}), \ldots ,F_{n}(x_{n})).
\end{equation*}
If $F_{i}(x_i)$ are continuous, then C is unique.
\end{thm}
Here, the functional $C$ is known as a copula and it gives
an alternative expression of the multivariate cumulative distribution
functions, i.e., it can be used for scale-free measures of 
dependencies on distributions such that 
$C(F_{1}, \ldots , F_{n})$ is the joint distribution
with the marginal distributions as the variables.
This means that the continuous multivariate marginal distributions
are able to be decomposed into $F_{1}, \ldots, F_{n}$
that expresses the behavior of random variables and
a copula $C$ that expresses the dependency structure of 
the random variables.
A copula with joint probability distributions,
in the case of two variables of Theorem \ref{thm:DEFINITION_OF_COPULA},
is sometimes called a `pair' copula.
A pair copula can be written as a graphical model.
Bedford and Cooke \textit{et al.} introduced a novel
technique that homologizes them\cite{Bedford2001}.
\subsection{Correlation}
A well-known indicator of variable dependencies is
the correlation coefficient. 
The correlation coefficient between the random variables
$X$ and $Y$ is defined as
$\rho = \frac{E((X-\mu_{1})(Y-\mu_{2}))}{\sigma_{1}\sigma_{2}}$,
where $\mu_{1}$ and $\mu_{2}$ are the means of $X$ and $Y$,
respectively, while $\sigma_{1}$ and $\sigma_{2}$ are the standard deviations of $X$ and $Y$, respectively.
If a joint probability distribution between two variables has
a correlation coefficient, a range of $\rho$ satisfies
$-1 \leq \rho \leq 1$.
The joint probability distribution for $X$ and $Y$ has a linear
correlation with an equation $y=a + bx$, $b > 0$, if $\rho = 1$.
This means that $P(Y=a+bX)=1$. Similarly, if $\rho = -1$,
we get the same result except $b < 0$.
If $\rho \neq \pm 1$, is there a line in the $xy$-plane
where the joint probability for $X$ and $Y$ is concentrated
with a width? 
Under certain constraints, this line actually exists. 
At this time, the $\rho$ can be regarded as a
measure of strength which the joint probability for
$X$ and $Y$ are concentrated on
the line\cite{Hogg2012}.
Clearly, linear correlation cannot represent dependencies 
that are non-linear. Rank correlation can be used to evade
these kinds of problems that represent a dependency between
the variables. Rank correlation is not a value of each variable
itself but is a correlation based on a rank of each variable
according to some criteria.
In this study, we used the Kendall rank correlation
coefficient (Kendall's $\tau$) as the rank correlations.
The population version of Kendall's $\tau$ is defined
as the probability of the concordance minus the
probability of the distance:
\begin{align*}
\tau = &\quad
P(\textrm{sgn} \{(X_{1} - X_{2}) (Y_{1} - Y_{2}) \} = 1) \\
    & \hspace{+1cm} -P(\textrm{sgn} \{(X_{1} - X_{2}) (Y_{1} - Y_{2}) \} = -1)
\end{align*}
If a pair of $\textrm{sgn} \{(X_{1} - X_{2}) (Y_{1} - Y_{2}) \} = 1$,
then we call this relationship concordance and there is an
increasing relationship between the random variables $X$ and $Y$.
Otherwise, we call this relationship
disconcordance and there is a decreasing relationship
between the variables $X$ and $Y$.
Furthermore, a range of Kendall's $\tau $ is $-1 \leq \tau \leq 1$.
Kendall's $\tau$ can be expressed using a copula.
\begin{thm}\upshape\cite{Nelsen2010} \itshape
Let $\tau$ denote the difference between the probabilities
of concordance and disconcordance of $(X_{1}, Y_{1})$
and $(X_{2}, Y_{2})$, then
\begin{align*}
    \tau = 4 \int \int_{I^{2}} C_{2} (u, v) d C_{1}(u, v) - 1 
\end{align*}
where $C_{1}$ and $C_{2}$ are the copulas of $(X_{1}, Y_{1})$ and $(X_{2}, Y_{2})$, respectively. The unit square $I^{2}$ is the product $I \times I$ where $I = [0, 1]$.
\end{thm}
In the next section, we explain how to visualize a neural network by using Kendall's $\tau$.
%

\section{ALGORITHM}
\label{sec:ALGORITHM}
\label{sec:ALG}
\begin{figure}[!ht]
\tikzset{%
  every neuron/.style={
    circle,
    draw,
    minimum size=1cm
  },
  every path neuron/.style={
    circle,
    draw=none,
    minimum size=1cm,
    fill=red!30,
    thick,
    text=black
  },
  neuron missing/.style={
    draw=none, 
    scale=4,
    text height=0.333cm,
    execute at begin node=\color{black}$\vdots$
  },
}
\begin{center}
\scalebox{0.6}{
\begin{tikzpicture}[x=1.5cm, y=1.5cm, >=stealth]
\foreach \m/\l [count=\y] in {1,2,3,missing,4}
  \node [every neuron/.try, neuron \m/.try] (input-\m) at (0,2.0-\y) {};
  
\foreach \m [count=\y] in {1,2,3,missing,4}
  \node [every neuron/.try, neuron \m/.try ] (hidden1-\m) at (2.5,2.5-\y*1.25) {};
\foreach \m [count=\y] in {1,2,missing,3}
  \node [every neuron/.try, neuron \m/.try ] (hidden2-\m) at (4.5,2.0-\y*1.25) {};
\foreach \m [count=\y] in {1,2,missing,3}
  \node [every neuron/.try, neuron \m/.try ] (output-\m) at (7,1.5-\y) {};
%
\foreach \m/\l [count=\y] in {1}{
  \node [every path neuron/.try, neuron \m/.try] (input-\m) at (0,2.0-\y) {x\y};
  \node [every path neuron/.try, neuron \m/.try ] (hidden1-\m) at (2.5,2.5-\y*1.25) {h\y(N)};
  \node [every path neuron/.try, neuron \m/.try ] (hidden2-\m) at (4.5,2.0-\y*1.25) {h\y(N+1)};
  \node [every path neuron/.try, neuron \m/.try ] (output-\m) at (7,1.5-\y) {pred\y};
}
%
\foreach \i in {1,...,4}
  \foreach \j in {1,...,4}
    \draw [-] (input-\i) -- (hidden1-\j);
\foreach \i in {1,...,4}
  \foreach \j in {1,...,3}
    \draw [-] (hidden1-\i) -- (hidden2-\j);
\foreach \i in {1,...,3}
  \foreach \j in {1,...,3}
    \draw [-] (hidden2-\i) -- (output-\j);
%
\foreach \i in {1}
  \foreach \j in {1}{
    \draw [-, color=red, thick] (input-\i) -- (hidden1-\j);
    \draw [-, color=red, thick] (hidden1-\i) -- (hidden2-\j);
    \draw [-, color=red, thick] (hidden2-\i) -- (output-\j);
   }
%
\node [align=center, above, scale=1.5] at (0.0,2) {Input\\layer};
\node [align=center, above, scale=1.5] at (2.5,2) {Hidden \\N layer};
\node [align=center, above, scale=1.5] at (4.5,2) {Hidden \\(N+1) layer};
\node [align=center, above, scale=1.5] at (7.0,2) {Output \\layer};
%
\node[fill=white,scale=3,inner xsep=0pt,inner ysep=15mm] at ($(input-1)!.5!(hidden1-4)$) {$\dots$};
\node[fill=white,scale=3,inner xsep=0pt,inner ysep=15mm] at ($(hidden2-1)!.5!(output-3)$) {$\dots$};
%
\draw [red, thick] (5.0,-4.0) -- (5.5,-4.0);
\node[fill=white,scale=2,inner xsep=0pt,inner ysep=0mm, text=black] at (6.2,-4.0) {path};
\end{tikzpicture}
}
\caption{A Neural Network with Nodes and Edges}
\label{fig:copula_operation}
\end{center}
\end{figure}
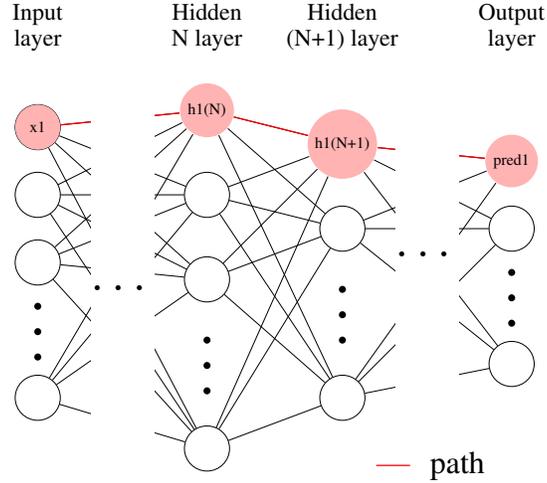

A neural network having nodes and edges can be drawn
as shown in Figure \ref{fig:copula_operation}.
Therefore, the correlation coefficients between the neural network
layers can be calculated by applying the concept of a pair
copula described in the previous section.
The paths are obtained by connecting the input layer
to the output layer in a forward direction.
An example of the paths is denoted as red nodes and edges.
The calculation result reveals which paths in a
neural network are important for the classification or prediction
calculation of the correlation coefficients. In this section,
we propose a new algorithm that makes the decision-making
of a neural network clear, known as the Copula-based
Visualization Technique (\textsc{CVT}) for a neural network, 
as shown in Algorithm \ref{alg:CVT}.

\begin{algorithm}[htb]
   \caption{\textsc{CVT}}
   \label{alg:CVT}
   \begin{enumerate}
     \item Initial Setting:
     \begin{enumerate}
     \item
     Train a neural network
     \item
     Compute all cumulative marginal distribution functions of any random variables
     \item
     Compute the correlation matrices between distributions of the activation function values of all nodes using training samples
     \item
     List all the paths. For example, see Figure \ref{fig:copula_operation}.
    \end{enumerate}
    \item
    Procedure:
    \begin{enumerate}
     \item
     Compute the Convolution of Correlation Coefficient (CCC).
     For example, CCC of red path in Figure \ref{fig:copula_operation}
     is $[x_{1}, \ldots, \mathrm{h}_{1}(N), \mathrm{h}_{1}(N+1), \ldots, \mathrm{pred1}]$.
     \item
     Compute the variance of CCC for a path with the same nodes and edges
     except the output nodes
    \end{enumerate}
   \end{enumerate}
\end{algorithm}

In the initial setting, we are required to train a neural network, the cumulative marginal distribution function, correlation matrix, and paths.
In the procedure step, we need to compute the variance of the convolution correlation coefficient: 
\begin{align*}
    \mathrm{VaR(CCC)}.
\end{align*}
This is the importance degree of each path for visualizing and understanding the neural network model. The variance of CCC expresses the path sensitivity.
For example, we can arrange the ranking in a descending order of $\mathrm{VaR(CCC)}$.
In general, it is not possible to understand how a model makes
a judgment based only on weights in a neural network\cite{Haufe2014}; \cite{Kindermans2018}.
For this reason, it is impossible to know how the feature value
affects the classification or prediction result and which node is
highly sensitive to the classification or prediction result
only by looking at the weights.

\section{VISUALIZATION OF A NEURAL NETWORK}
\label{sec:EVALUATION}
\begin{figure}[!ht]
    \begin{center}
    \includegraphics[clip, scale=0.29]{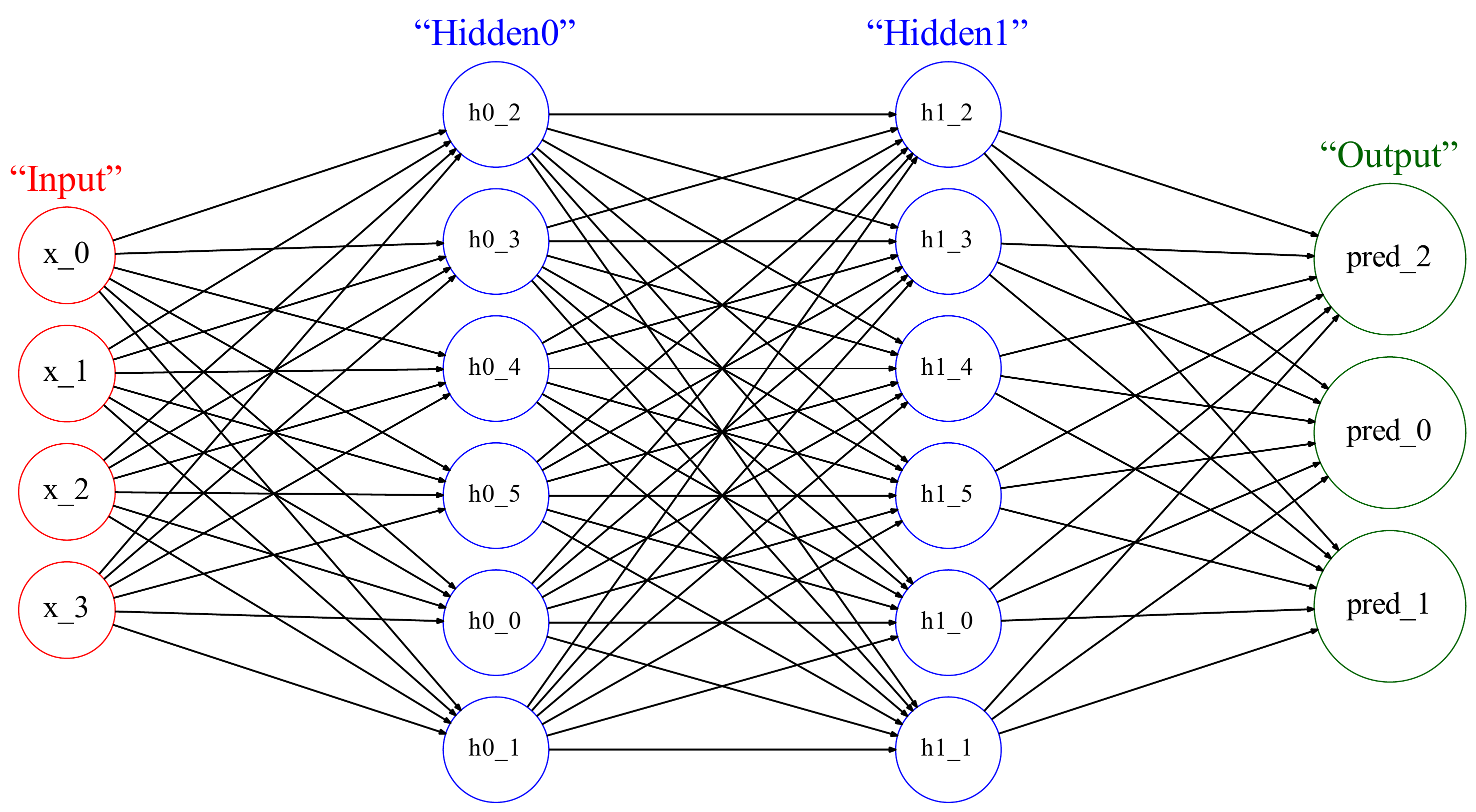}
    \caption{Structure of Trained Neural Network}
    \label{NNmodel}
    \end{center}
\end{figure}

The experiment with \textsc{CVT} to a neural network is described
in this section. Subsection \ref{sec:EXPERIMENTAL_SET_UP}
presents the experimental set up such as the dataset and 
the calculation of the probability density functions (PDFs) 
and cumulative distribution functions (CDFs).
Subsection \ref{sec:CCAL} details the simultaneous 
visualization of the important feature values
and the paths that are traced in the process of decision-making.
This result presents our main contribution.
The feature values obtained by \textsc{CVT} were ranked
according to VaR(CCC) and were compared with the
Random Forest result in Subsection \ref{sec:COMPARISON}.
All the source codes to demonstrate this experiment are
available at Github\footnote{\url{https://github.com/covit2019/analysis\_codes}}.
\label{sec:EXPERIMENTAL_SET_UP}
The Fisher's Iris data set\footnote{\url{http://archive.ics.uci.edu/ml/datasets/Iris}} included in the scikit-learn package\footnote{\url{https://scikit-learn.org/stable/auto\_examples/datasets/\\plot\_iris\_dataset.html}}
was used to train the neural network.
The values of $\it{sepal}\,\it{length}$, $\it{sepal}\,\it{width}$,
$\it{petal}\,\it{length}$, and  $\it{petal}\,\it{width}$ were used
as the input feature values which defined
$\mathrm{x}\_0$, $\mathrm{x}\_1$, $\mathrm{x}\_2$ and $\mathrm{x}\_3$, respectively.
The labels of $\it{Setosa}$, $\it{Versicolor}$, $\it{Virginica}$ were used as
one-hot encoding vector.
Here, $\it{Setosa}$ is $\mathrm{pred}\_0$, $\it{Versicolor}$ $\mathrm{pred}\_1$, $\it{Virginica}$ $\mathrm{pred}\_2$.
The structure of the trained neural
network is illustrated in Figure \ref{NNmodel}, comprised of 
four nodes (red) in the input layer, six (blue) in the first hidden layer, six (blue) in the second hidden layers, and three (green) in the output layer.
All the nodes in the hidden layers and the output layer express the values 
through the activation functions.
Rectified Linear Unit (ReLU) was used as the activation function 
in the hidden layers and Softmax was used in the output function.
This model is a standard neural network model.
The data used for training was 120 samples out of 150 and the
accuracy rate was approximately $97.5\%$.
\begin{figure}[!ht]
    \begin{center}
	\hspace*{-5mm}
    \includegraphics[clip, scale=0.45]{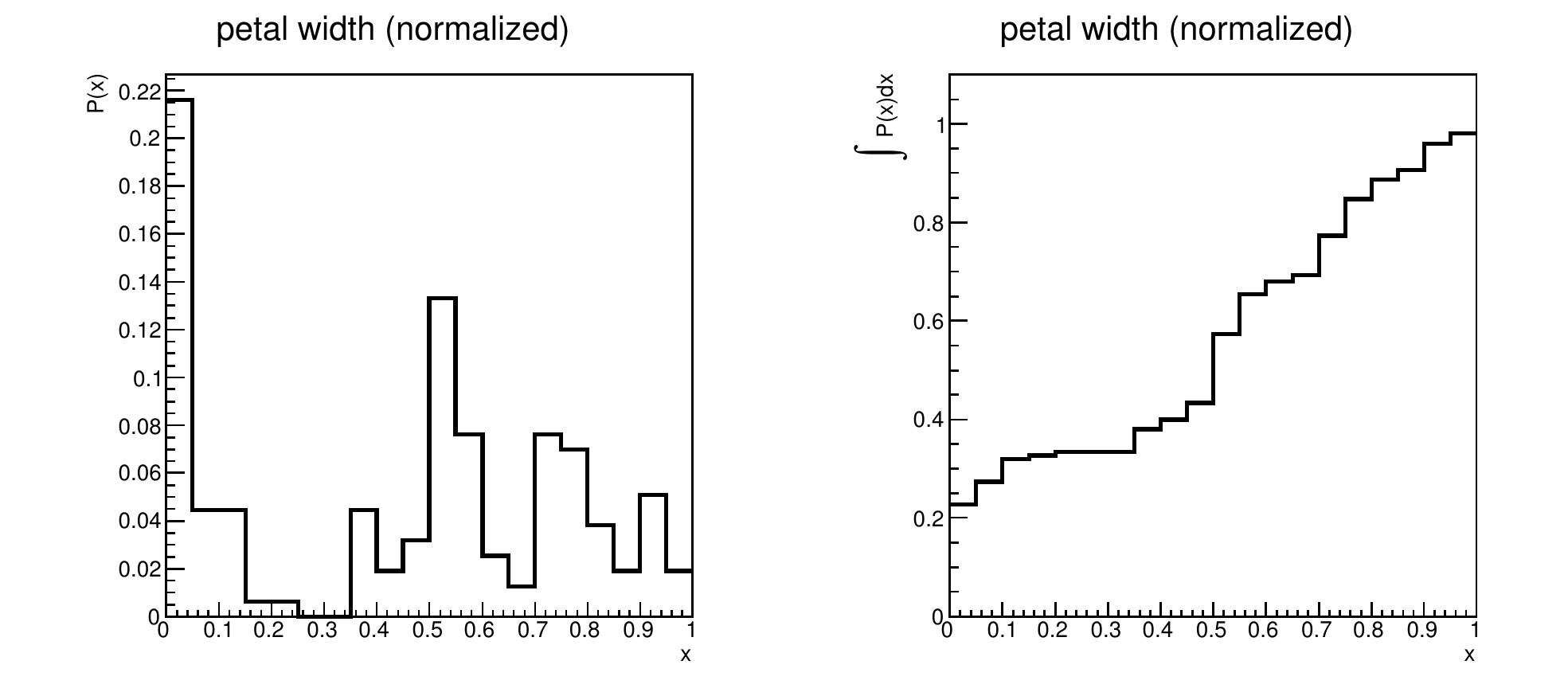}
    \caption{Example of Conversion from a PDF (left)
    to a CDF (right) at Each Node}
    \label{PDFtoCDF}
    \end{center}
\end{figure}

CDFs are required to calculate the correlation coefficients as
described in Section \ref{sec:COPULA}.
Therefore, we first created histograms of the outputs of
the nodes at the time of prediction.
These histograms are equivalent to PDFs as shown 
in the left-side of Figure \ref{PDFtoCDF}.
Here, the PDFs at the input nodes were, in fact, the feature values. 
Next, we integrated them to obtain the CDFs
as shown in the right-side of Figure \ref{PDFtoCDF}.
The $Y$-axis values of the CDFs histograms were used for
this calculation.
\subsection{Calculation of Correlation Coefficients
            Between Neural Network Layers}
\label{sec:CCAL}
\begin{figure}[!ht]
    \begin{center}
    \includegraphics[clip, scale=0.29]{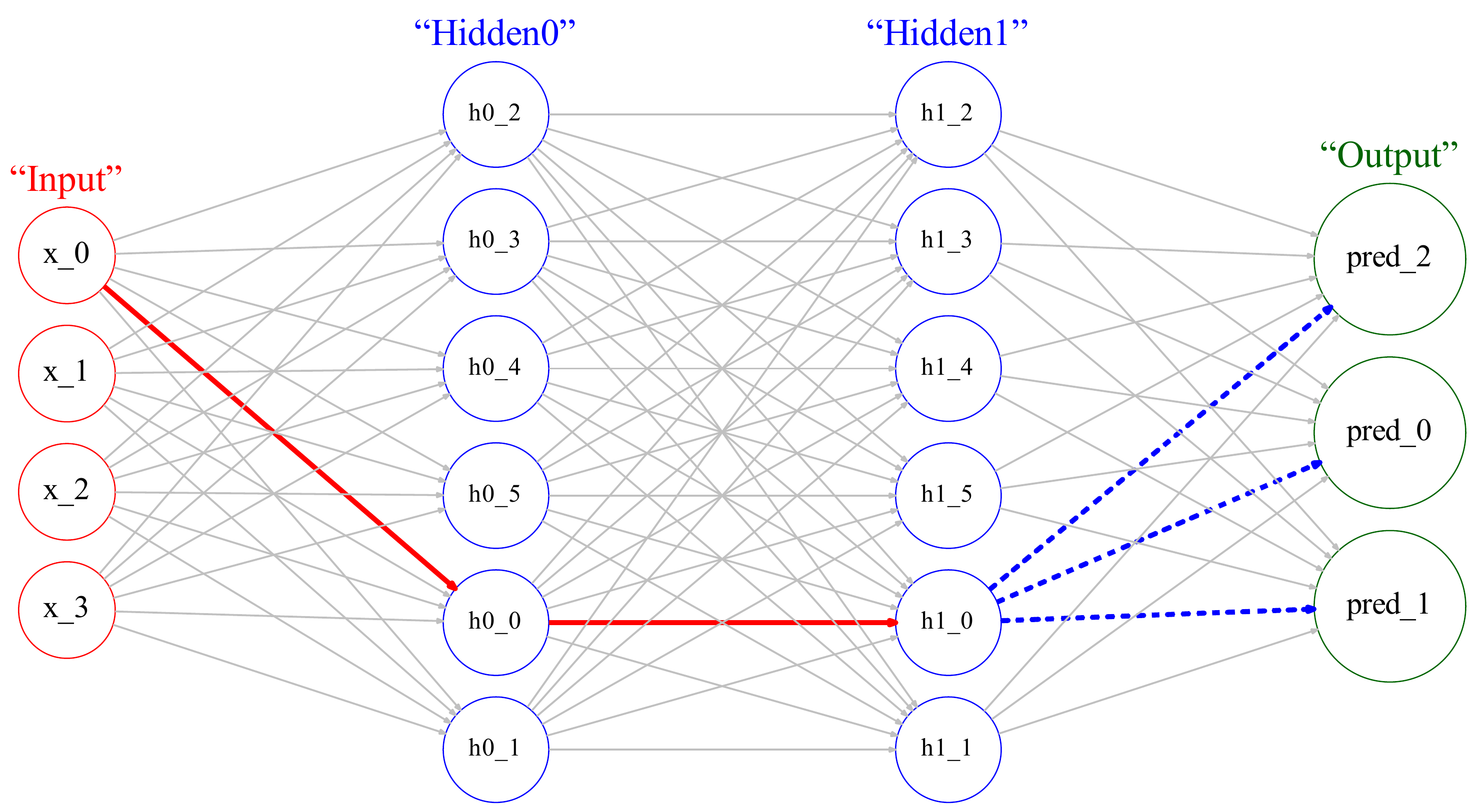}
    \caption{Example of Paths from Input Nodes to Output Nodes}
    \label{root}
    \end{center}
\end{figure}
As shown in Figure \ref{root}, the number of paths from the input
to the second hidden layer (Hidden1)
in the neural network is $4 \times 6\times 6 = 144$ and
each path is branched off to the three outputs.
An example path from the 144 paths is drawn as a red line 
and the branch lines are drawn as blue dotted lines.
Next, $\mathrm{VaR(CCC)}$ for each path was calculated using the \textsc{CVT} algorithm described in Section \ref{sec:ALG}. The calculation of the correlation coefficients was executed by the noble C++ library: vinecopulib\cite{Nagler2019}.
\begin{table*}[htb]
    \caption{Top 10 Important Path Ranking}
    \begin{center}
        \begin{tabular}{|c||c|c|c||c||c|} \hline
            Path & Setosa & Versicolour & Virginica & VaR(CCC) \\ \hline \hline
            $\mathrm{x}\_2$, $\mathrm{h}0\_5$, $\mathrm{h}1\_1$ & -0.00801  & 0.047293 & 0.69013 & 0.150615 \\
            $\mathrm{x}\_2$, $\mathrm{h}0\_3$, $\mathrm{h}1\_1$ & -0.00801  & 0.047293 & 0.69013 & 0.150615 \\
            $\mathrm{x}\_3$, $\mathrm{h}0\_3$, $\mathrm{h}1\_1$ & -0.00801  & 0.047293 & 0.69013 & 0.150615 \\
            $\mathrm{x}\_3$, $\mathrm{h}0\_5$, $\mathrm{h}1\_1$ & -0.00801  & 0.047293 & 0.69013 & 0.150615 \\
            $\mathrm{x}\_2$, $\mathrm{h}0\_5$, $\mathrm{h}1\_5$ & -0.00551  & 0.047878 & 0.69013 & 0.149875 \\
            $\mathrm{x}\_2$, $\mathrm{h}0\_3$, $\mathrm{h}1\_5$ & -0.00551  & 0.047878 & 0.69013 & 0.149875 \\
            $\mathrm{x}\_3$, $\mathrm{h}0\_5$, $\mathrm{h}1\_5$ & -0.00551  & 0.047878 & 0.69013 & 0.149875 \\
            $\mathrm{x}\_3$, $\mathrm{h}0\_3$, $\mathrm{h}1\_5$ & -0.00551  & 0.047878 & 0.69013 & 0.149875 \\
            $\mathrm{x}\_2$, $\mathrm{h}0\_3$, $\mathrm{h}1\_4$ & 0.0071334 & 0.074663 & 0.69013 & 0.126953 \\
            $\mathrm{x}\_2$, $\mathrm{h}0\_5$, $\mathrm{h}1\_4$ & 0.0071334 & 0.074663 & 0.69013 & 0.126953 \\ \hline
        \end{tabular}
        \label{tb:CopulaRankTable}
    \end{center}
\end{table*}
\begin{figure*}[!ht]
	\begin{center}
        \includegraphics[clip, scale=0.5]{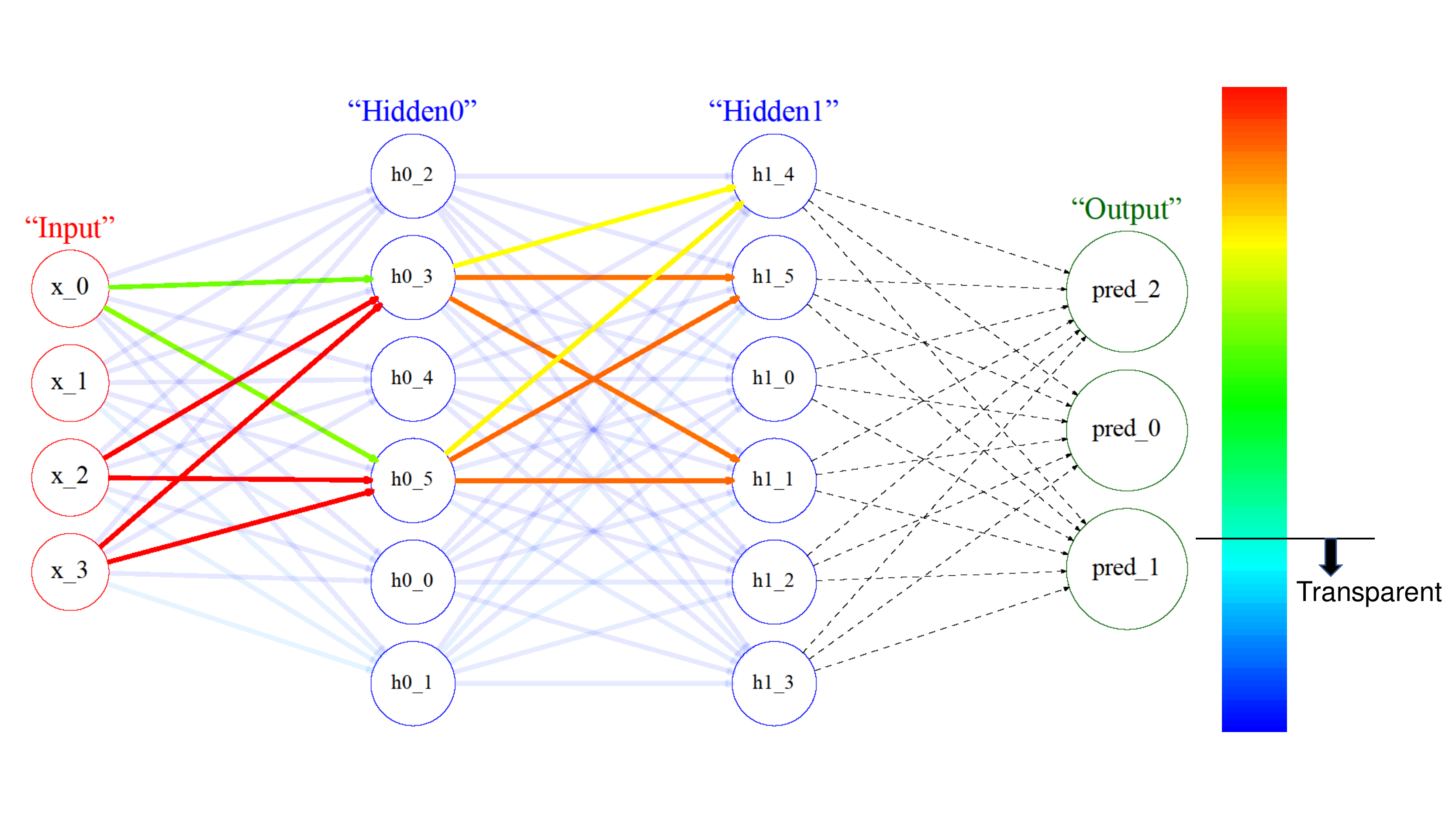}
        \caption{Visualization of the Neural Network Decision-Making}
        \label{CopulaRank}
    \end{center}
\end{figure*}
Table \ref{tb:CopulaRankTable} presents the top 10 important path
ranking in a descending order of $\mathrm{VaR(CCC)}$. This table presents which
paths are important for decision-making of the neural network
from which the feature values $\mathrm{x}\_2$ and $\mathrm{x}\_3$
are especially important. Besides that, Figure \ref{CopulaRank} shows
the visualization of all the paths. The importance defined as
the sum of $\mathrm{VaR(CCC)}$ of paths through the edge is portrayed as a contrasting density, and the density corresponding
to blue (less important) is set transparent to emphasize the important paths.
The values listed in Table \ref{tb:CopulaRankTable} are sometimes
the same because the CDFs have discrete histograms.
This problem can be avoided by making the bin size of
the histograms finer or by using CDFs, which are estimated 
through function fitting.
\subsection{Comparison with Random Forest}
\label{sec:COMPARISON}
\begin{figure}[!htb]
    \begin{center}
	\hspace*{-5mm}
    \includegraphics[clip, scale=0.32]{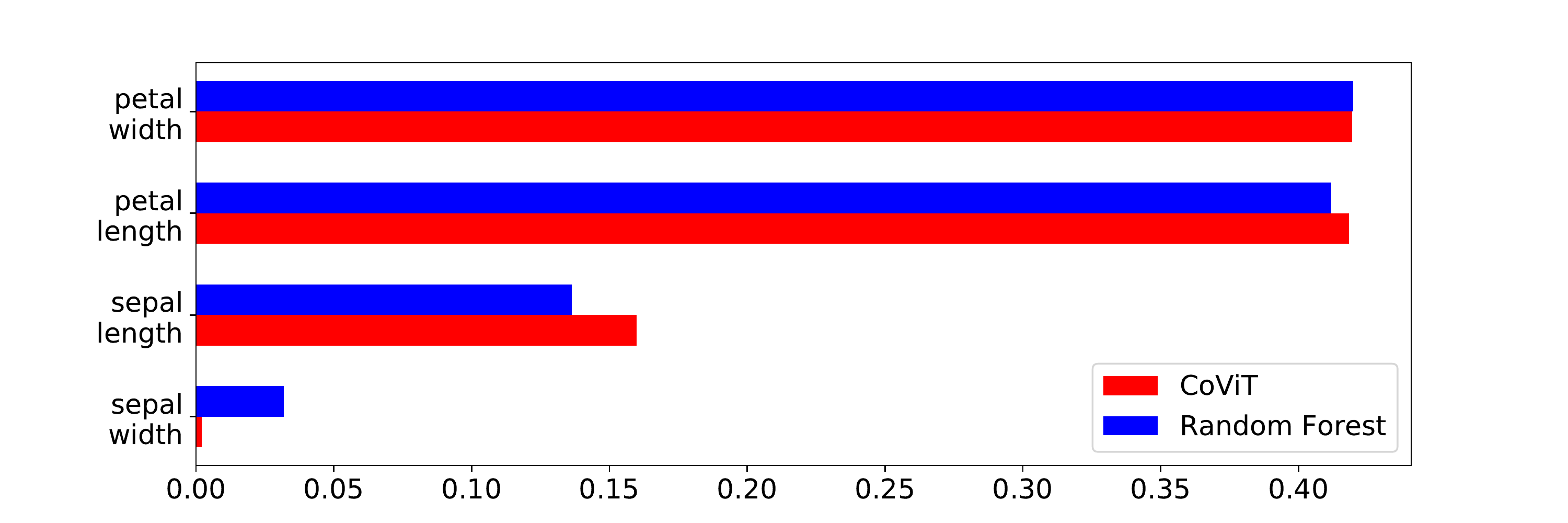}
    \caption{Comparison of Feature Importance with Random Forest}
    \label{comparison_with_RF}
    \end{center}
\end{figure}
We compared the \textsc{CVT} results with those obtained by 
Random Forest for the validity of interpretability. 
Random Forest implemented in the scikit-learn package\footnote{\url{https://scikit-learn.org/stable/modules/generated/sklearn.ensemble.RandomForestClassifier.html}}
with the default parameters were used.
The data used for training had 120 samples out of 150 and
the accuracy rate was approximately $97\%$. This accuracy rate
was extremely close to that of the neural network experimented 
in the previous subsection.
Random Forest, which is sometimes regarded as one of the 
highly interpretable algorithms, can calculate the importance 
of the feature values. 
Therefore, we compared the Random Forest feature
importance values with the following:
\begin{equation*}
    E^{x_{i}}(\mathrm{VaR(CCC)})
\end{equation*}
where $E^{x_{i}}( \cdot )$ is an operator that takes the expected
value with respect to the feature values $x_{i}$ and normalized
to be compared.
For example, $E^{x_{2}}(\mathrm{VaR(CCC)})$ means that the average of all paths begin with $\mathrm{x}\_2$, i.e., that of the first row, the second, the fifth, ..., and so on as shown in Table \ref{tb:CopulaRankTable}.
Figure \ref{comparison_with_RF} shows the result.
It was not compared directly because the importance was calculated 
using different methods, but the results were consistent with each other. This means that \textsc{CVT} can derive the importance
of feature values in a neural network.
%

\section{RELATED WORKS}
As summarized in the introduction, sensitivity analysis method\cite{Zeiler2014}; \cite{Smilkov2017} 
and reverse tracing method\cite{Bach2015} have been studied intensively in the recent years.
\textsc{CVT} is complementary to the aforementioned methods in terms of explaining the models as these methods do not address the problem of explaining
important feature values and paths that contribute to the results of classification or prediction through all training data.
In other words, \textsc{CVT} is not appropriate for measuring the effect and contribution of feature values
of individuals
when only a portion of the data is focused on.
It is difficult to say which of these methods is better, and therefore, an appropriate method should be chosen based on the situation.

\section{CONCLUSION AND FUTURE WORKS}
\label{sec:CONCLUSION}
In this study, we proposed a Copula-based Visualization Technique (\textsc{CVT})
for a neural network.
\textsc{CVT} can easily visualize the decision-making process of a neural network
using correlation coefficients.
Information regarding the feature values that are considered to be important by the trained neural
network and paths that are mainly
traced in the process of decision-making are obtained by the proposed
algorithm. The experimental result of the proposed algorithm was consistent with
the importance values estimated using Random Forest.
This suggests that \textsc{CVT} can be used for the
interpretability of a neural network.

\textsc{CVT} was experimented using a simple neural network but can also be extended to other
algorithms, including a deep neural network that can be considered as a graphical model.
\textsc{CVT} can serve as a general method for machine
learning interpretability. Kendall's $\tau$ was used as the
correlation coefficients, but other correlation
coefficients, by all means, can also be used.
Future works include the theoretical analysis and the use of other correlation coefficients for \textsc{CVT}.
Additionally, \textsc{CVT} can suggest
an approach for compressing a neural network as doing so helps in identifying the nodes and edges that contribute to the classification or
prediction results. \textsc{CVT} can also use parameter
tuning for the abovementioned reason. Additionally, \textsc{CVT} may be able
to reveal the reasons for Adversarial
Attack cheating a neural network,
which can also be considered for future work.

%
%

\clearpage

\bibliographystyle{unsrt}  


\end{document}